\documentclass[10pt, a4paper]{article}
\usepackage{lrec-coling2024} 

\usepackage{amsmath,amsfonts,amssymb}
\usepackage{multirow}
\usepackage{array}
\usepackage{arydshln}
\usepackage{booktabs}
\usepackage{tabularx}
\usepackage{caption}
\usepackage{listings}
\usepackage{bm}
\usepackage{xcolor}  

\usepackage{xcolor}
\usepackage{hyperref}
 \definecolor{darkblue}{rgb}{0, 0, 0.5}
  \hypersetup{colorlinks=true, citecolor=darkblue, linkcolor=darkblue, urlcolor=darkblue}

\usepackage{xstring}

\usepackage{color}

\title{Semi-Instruct: Bridging Natural-Instruct and Self-Instruct for Code Large Language Models}

\name{Xianzhen Luo$^1$, Qingfu Zhu$^{1*}$\thanks{\ \ * Corresponding author}, Zhiming Zhang$^1$, \\ 
\textbf{Xu Wang}$^2$, \textbf{Qing Yang}$^2$, \textbf{Dongliang Xu}$^2$, \textbf{Wanxiang Che}$^1$} 

\address{
$^1$Harbin Institute of Technology, Harbin, China,\\
$^2$Du Xiaoman (Beijing) Science Technology Co., Ltd.,\\
\texttt{\{xzluo, qfzhu, zmzhang, car\}@ir.hit.edu.cn}, \\
\texttt{\{wangxu04, yangqing, xudongliang\}@duxiaoman.com}}

\abstract{
Instruction tuning plays a pivotal role in Code Large Language Models (Code LLMs) for the task of program synthesis. Presently, two dominant paradigms for collecting tuning data are natural-instruct (human-written) and self-instruct (automatically generated).
Natural-instruct includes diverse and correct codes but lacks instruction-code pairs, and exists improper code formats like nested single-line codes. In contrast, self-instruct automatically generates proper paired data. However, it suffers from low diversity due to generating duplicates and cannot ensure the correctness of codes. 
To bridge the both paradigms, we propose \textbf{Semi-Instruct}.
It first converts diverse but improper codes from natural-instruct into proper instruction-code pairs through a method similar to self-instruct. To verify the correctness of generated codes, we design a novel way to construct test cases by generating cases' inputs and executing correct codes from natural-instruct to get outputs. Finally, diverse and correct instruction-code pairs are retained for instruction tuning. 
Experiments show that semi-instruct is significantly better than natural-instruct and self-instruct. 
Furthermore, the performance steadily improves as data scale increases. Our code and data are public at [link].
 \\ \newline \Keywords{Program Synthesis, Large Language Model, Self-Instruct} }

\begin{document}



\maketitleabstract

\section{Introduction}
Program synthesis aims to generate code snippets given a specification, typically framed as a natural language description~\citep{prosyn}. It can effectively enhance programming efficiency and improve productivity. Meanwhile, the coding ability has been observed to be positively correlated with the performance of large language models (LLMs) on reasoning tasks~\citep{shin-van-durme-2022-shot, yang-etal-2022-generating, chen2022program}, which further increases the attention to program synthesis.
Similar to other generative tasks, the common practice of enhancing program synthesis ability is fine-tuning code LLMs on instruction-code pairs to align with human intentions \citep{codet5+,pangucoder2,octopack}.
According to the source of instruction tuning data, the approaches for data collection can be further divided into two categories: \textbf{Natural-Instruct} (NI)~\citep{mishra-etal-2022-cross,alphacode} and \textbf{Self-Instruct} (SI)~\citep{self-instruct,wizardcoder}.

\begin{figure*}[!ht]
\begin{center}
\includegraphics[width=\textwidth]{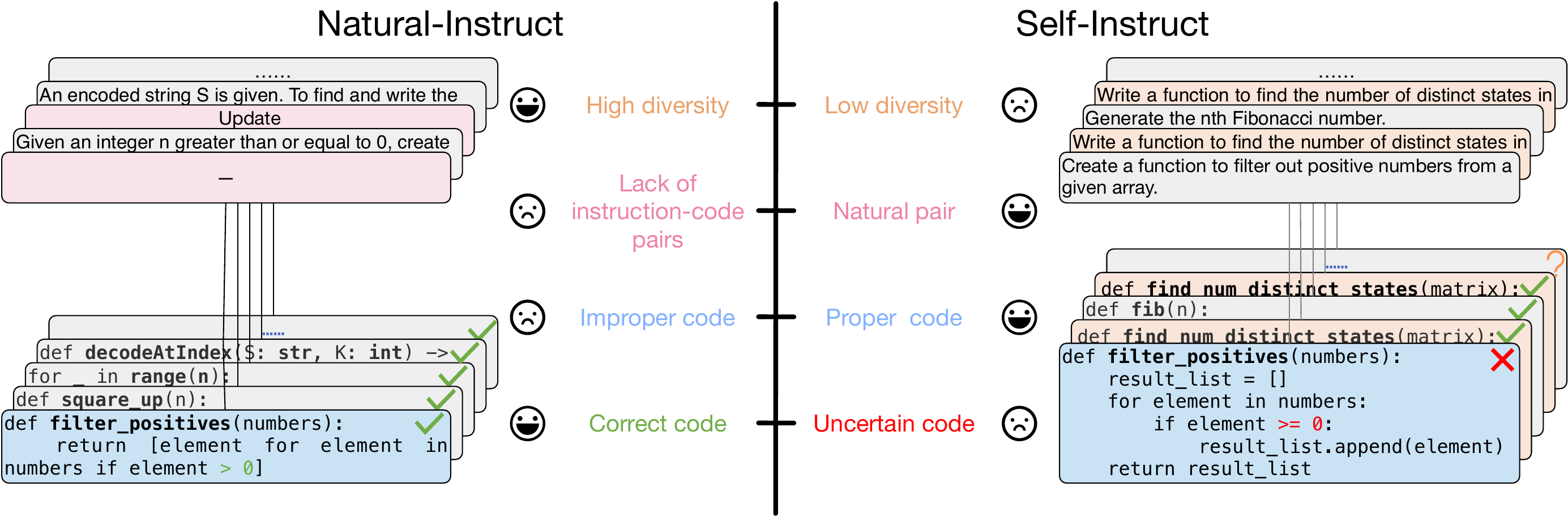} 
\caption{The advantages and disadvantages of natural-instruct and self-instruct. Their colors correspond to the data samples.}
\vspace{-1em}
\label{fig.1}
\end{center}
\end{figure*}

NI aims at collecting human-written data from many code-related platforms such as GitHub and Codeforces. It consists of natural language such as extracted program comments or problem descriptions and corresponding codes or solutions. Two advantages lie in its \textbf{\textit{diverse}} and \textbf{\textit{correct}}~\citep{zan-etal-2023-large} codes, as shown in the left part of Fig~\ref{fig.1}. On the one hand, such platforms offer massive diverse codes with distinct functionalities. On the other hand, the correctness of codes can be guaranteed by carefully controlling data sources such as selecting solutions from contest websites that have passed all human-created test cases.
However, two drawbacks limit the performance of code LLMs tuning on NI data. First, \textbf{\textit{improper}} coding formats such as nested single-line code~\citep{10.1145/1985362.1985365}, and ambiguous variable names add too much noise to the data.  
Second, since natural languages always count only a small percentage (6.67\%)~\citep{plbart} and most of them (sometimes formed as a single word such as ``update") can not serve as instructions, the \textbf{\textit{lack of instruction-code pairs}} which are high-quality and complete limits the program synthesis ability of code LLMs gained from instruction learning. 
Therefore, while NI provides massive diverse, and correct data, it is plagued by issues of improper coding formats and missing instruction-code pairs.

Conversely, SI leverages LLMs to automatically generate \textbf{\textit{naturally paired}}~\citep{alpaca} instructions and codes with \textbf{\textit{proper}} format without human effort. 
However, its two shortcomings can not be ignored. First, the generated instruction-code pairs are of \textit{\textbf{low diversity}}~\citep{self-instruct} due to the constrained number of seed prompts. Second, absent from any testing or manual calibration, the correctness of generated codes is \textbf{\textit{uncertain}}~\citep{siren}.
Some approaches attempt to validate code by generating test cases consisting of inputs and outputs\citep{codet, codellama}. While the generated inputs are mostly correct, we often fail to obtain expected outputs when faced with complex scenarios such as logical and numerical reasoning. Uncertain test cases not only waste generation resources but also are challenging to filter out. 
Thus, while SI excels in generating complete and clear instruction-code pairs, its limitations lie in the repetitive nature and inability to ensure correctness.



To this end,  we introduce a novel approach, \textbf{Semi-Instruct} (SemI), to bridge the inherent strengths of both NI and SI.
By leveraging the generative capability of LLMs like what SI does and feeding the diverse codes from NI (original code) into them, we can obtain the naturally paired instructions and proper codes (refined codes) that are correspondingly of high diversity. 
To validate the correctness of such refined codes, instead of directly generating the complete test cases at once as previous methods do, we offer a novel solution to handle it. Specifically, recall that it can always be guaranteed to obtain the correct inputs of test cases from LLMs, the corresponding correct outputs are supposed to be easily derived by executing the original codes, resulting in the more reliable test cases. Thus, these determined test cases function to affirm the accuracy of the refined codes.

The detailed process of SemI is divided into three steps: 
(1) \textbf{Generation}:
Starting with an original code, LLMs generate an instruction, a fixed number of test cases' inputs based on the instruction, and a refined code in sequence.
(2) \textbf{Validation}: 
Initially, we obtain outputs of test cases by running the generated inputs through the original code. However, since some inputs may cause runtime errors, not all of them result in complete test cases after execution. Then only the refined code that passes the remaining test cases is left. Finally, we eliminate any instructions that closely resemble previously generated ones.
(3) \textbf{Ranking}: Intuitively, the quality of generated input depends on to what extent the LLMs understand the instruction. The more difficult the instruction is, the fewer test cases can be constructed after execution. Inspired by curriculum learning \citep{cl}, data are organized in descending order based on the count of test cases before tunning. 
\textbf{To the best of our knowledge, we are the first to get test cases consisting of generating inputs and executing outputs, and use the amount as a measure of difficulty.}
We carry out extensive experiments on the widely-used HumanEval dataset \citep{codex}. 
When using only one type of data, SemI largely outperforms NI on each scale of data size and is also better than SI. Moreover, combining the data from SI and SemI outperforms SI alone by an average of 3\% on p@1. Most importantly, the performance keeps steadily rising instead of oscillating or declining with the amount of data scale increasing.

The contributions are listed as follows:
\begin{itemize}
    \item{We propose a novel method named \textbf{Semi-Instruct}, which bridges the natural-instruct and self-instruct. Through semi-instruct, we can obtain diverse and correct instruction-code pairs for instruction tuning code LLMs to improve the ability of program synthesis.}
    \item{Through executing on original code, we generate test cases in a more effective way. In addition, we offer a new perspective on using them as a measure of instruction difficulty.}
    \item{After adding semi-instruct data to self-instruct, the performance is better than only increasing self-instruct data. And the combination breaks out the self-instruct's dilemma and enables performance to grow as data increases. }
\end{itemize}

\begin{figure*}[!ht]
\begin{center}
\includegraphics[width=\textwidth]{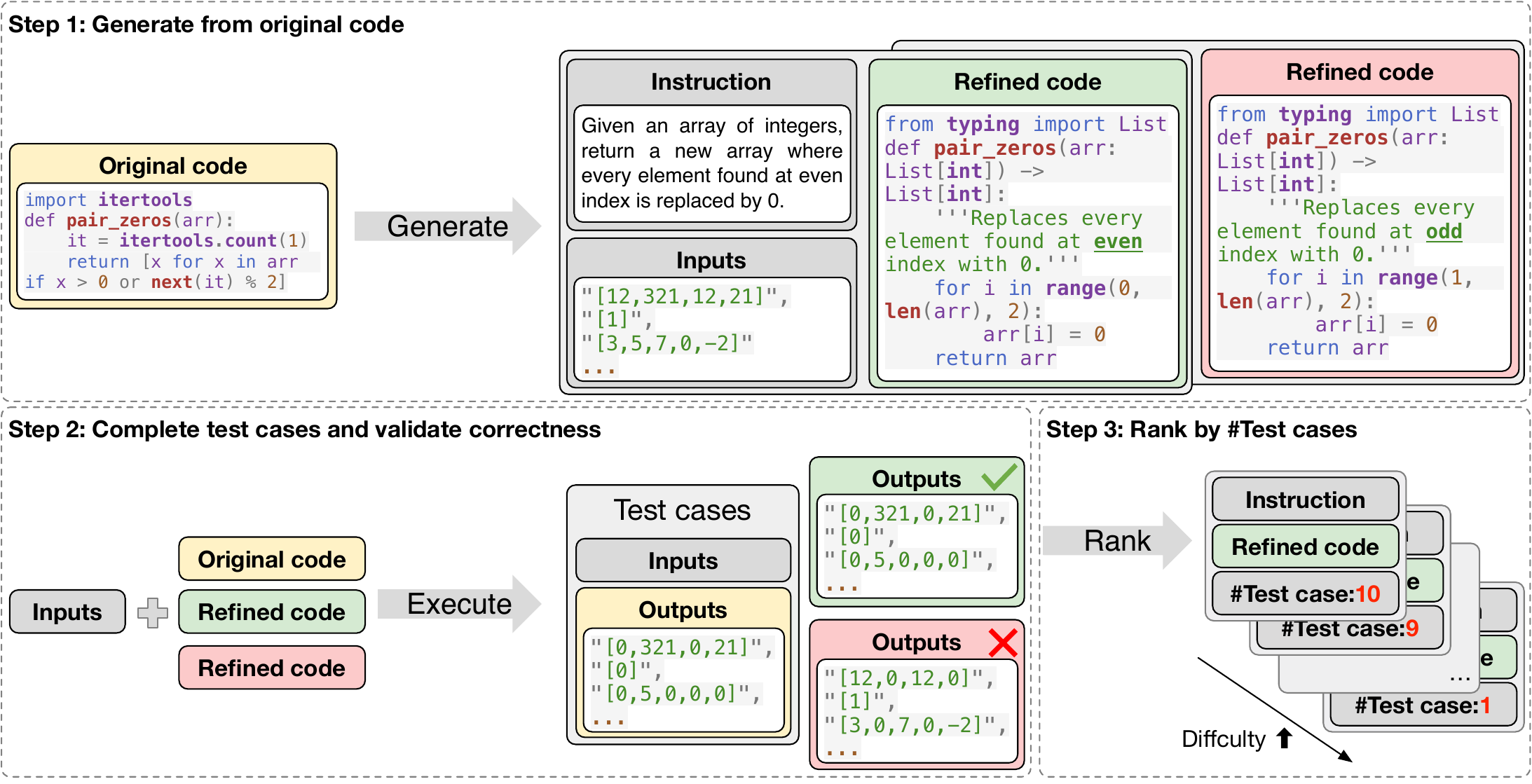} 
\caption{Pipeline of Semi-Instruct. It has three main components. \textbf{(1) Generation}: Given the original codes, generate instructions, a fixed number of test cases' inputs based on the instructions, and refined codes; \textbf{(2) Validation}: Run the original codes on the inputs, obtain complete test cases through extracting outputs from those successful executions, and subsequently retain refined codes that pass all test cases; \textbf{(3) Ranking}: Since the more difficult the instructions are, the less test cases are constructed, sort the data in reverse order according to the number of test cases.}
\vspace{-1em}
\label{fig.2}
\end{center}
\end{figure*}

\section{Related Work}
\subsection{Code LLM}
Due to the poor performance of general LLMs on program synthesis, some work~\citeplanguageresource{thestack,polycoder} collect large-scale code data from GitHub, and so on to train specific code LLMs. Pure code LLMs~\citep{alphacode, incoder, starcoder,santacoder} only pre-train on code data from scratch. Others~\citep{codex, nijkamp2023codegen, codellama} use the code data to secondary-pre-train. Although syntax errors appear fewer and fewer in the generated codes~\citep{self-edit}, they are limited to generating code based on natural language. 

\subsection{Code Instruction Tuning}
To improve the code LLMs' ability on program synthesis, many works construct aligned instruction-code pairs that match human intentions to finetune the LLMs. The two main types of these data are \textbf{natural-instruct} and \textbf{self-instruct}. 

NI consists of human-written text and code collected from
open-source platforms. CodeSearchNet~\citeplanguageresource{husain2019codesearchnet} collected publicly available GitHub repositories and extracted comments as instructions. Others~\citeplanguageresource{apps, alphacode, codenet} are derived from online judge websites such as Leetcode, Codeforces, and so on. Problem descriptions are seen as instructions. NI is used by earlier code LLMs~\citep{alphacode, coderl, codex, wang-etal-2021-codet5}. Codes in NI are diverse and correct. However, NI suffers from improper code formats and a lack of high-quality instruction-code pairs.

SI is inspired by Alpaca~\citep{alpaca} and first implemented by Code Alpaca~\citep{codealpaca}. phi-1~\citep{phi} used GPT-4 to generate a textbook-level high-quality corpus, while WizardCoder~\citep{wizardcoder} and Pangu-Coder2~\citep{pangucoder2} used the evol-instruct~\citep{wizardlm} to extend the large data from the basic Code Alpaca. SI has naturally instruction-code paired data and its codes are proper. However, it has low diversity due to generating duplicates and generated codes are uncertain.

\subsection{Test Cases}
Unlike natural languages, programs can only be considered correct if pass all the test cases written by developers~\citep{apps}. Many works proposed to generate test cases automatically to reduce human effort. Earlier heuristic methods based on search\citep{mio,pynguin} have limitations in diversity and quantity. Later works~\citep{alphacode, tufano2021unit} finetuned pretrained language models on existing labeled data to generate new test cases. Recent works~\citep{codet,codellama} utilize LLMs to sample without training. However, all of them can't guarantee the correctness of test cases

NI and SI are used separately in all previous work. We combine the two approaches to solve each other's problems. And we propose a novel way to construct test cases effectively and offer a new perspective on using them.


\section{Methodology}
\label{method}
\textbf{Semi-Instruct} takes advantage of both natural-instruct and self-instruct.
It generates an instruction given an original code to obtain aligned pairs similar to self-instruct, and maintains the diversity and functional correctness of natural-instruct at the same time.
Concretely, semi-instruct includes three steps, as shown in Figure~\ref{fig.2}.
First, given an original code, it generates an instruction, a refined code and test case inputs via LLM.
Second, verify the diversity of the instruction by its similarity to other instructions, and verify the functional correctness of the refined code on test cases.
Note the output of the test case is obtained by executing the input of the test case (generated in the last step) on the original code.
Finally, rank the pairs (consist of the instruction and refined code) in descending order based on the difficulty.
Here, the difficulty is measured by the number of test cases that the refined code passes.
We denote the three steps as generation, validation, and ranking, respectively.

\subsection{Generation}
The generation phase is not only about converting one-sided data into paired data, but also generate some auxiliary information to verify the quality later.

To collect pairwise data, a corresponding instruction needs to be generated first from the original code. Take inspiration from self-instruct that utilizes LLMs to generate instructions and codes at the same time, the original code can be refined during the generation process to solve the improper code format. This includes expanding nested single-line codes, renaming variables, adding necessary comments, etc. LLMs can generate the refined code with the paired instruction simultaneously, after comprehensively understanding the original code.

Test cases are essential to verify the consistency of the instruction and the correctness of the refined code. The common practice is generating whole test cases consisting of inputs and outputs through LLMs. While most inputs align with the instruction's constraints, LLMs frequently struggle to produce accurate outputs, especially when the instruction describes a complex task that needs logical and numerical reasoning. These inaccurate outputs consume significant generation resources and are challenging to filter out. Therefore, we only generate fixed number of inputs, and introducing a novel approach for test cases construction without generating outputs, described in Sec~\ref{validation}.

To use test cases, the answer type needs to be identified, which dictates how the inputs passing into code and how the code return outputs. LLMs can analyze the code and determine its type. If it's call-based, the function name will be extracted. A clear description of each component is below:



\textbf{Instruction}: A clear natural language description used as instruction in the tuning stage. It should directly reflect the function of the original code without too many implement details. 

\textbf{Refined Code}: A refined version of the original code used in the tuning stage. It should fix the improper format of the original code, without changing the behavior of the original function. 

\textbf{Answer Type}: The way of passing parameters into the original code. It should only be "Call-Based" (receiving input from parameters) or "Standard Input" (reading input from the standard input). If the answer type is "Call-Based", the function name should be included in test cases.

\textbf{Test Cases}: Test cases to validate the refined code based on the requirements of the instruction. Only inputs are generated without outputs during this stage. The existence of the function name is dependent on the answer type. 

After adding the original code into the prompt containing definitions of each component with a few examples, we feed the concatenated context into LLMs and extracted each component from the output. The unified prompt template is shown in the appendix. Through the generation stage, the diverse and improper code from NI become proper code with paired instructions. The generations are correspondingly of high diversity. The determined answer type and inputs of test cases will subsequently facilitate the validation phase.

\subsection{Validation}
\label{validation}
Limited by the capabilities of the LLMs, previously generated data needs to be further verified. We propose a novel way to construct completed test cases more effectively and filter out matching, correct, and diverse data in this stage. 

First, construct complete test cases by executing the original code on all inputs and gathering the corresponding outputs. Unlike generated outputs, executed outputs are inherently correct due to the correctness of original code. However, some inputs may report runtime errors during execution and result in no outputs. In addition to errors of the inputs themselves, the reason could be a mismatch between instruction and the original code, since the inputs are generated based on instruction. We directly discard the data that all inputs result in no outputs. The remaining data have different numbers of test cases but at least one. Our approach of constructing test cases saves the generation resources and ensures correctness.

Second, check the correctness of the refined code. LLMs naturally cannot ensure the generated code is correct. But the correctness of the original code can be used to verify the refined codes, by using the complete test cases constructed earlier. If there are inputs that report errors when the refined code is running, or if the results don't match the gold outputs, we assume that the refined code doesn't keep functional consistency with the original code. Only when all the cases pass, the refined code is considered correct and retained. 


Last, too similar data are removed. Original codes used to solve the same problem have the potential to generate very similar instructions, which can confuse the model. But this kind of data can also help the model generate more diverse codes. Therefore, we perform a looser filtering based on the ROUGE-L score of the instruction. refined code diversity is inherited from the original code, and the instruction diversity is guaranteed by the filtering.

We use the correctness of the original code to construct test cases and then pick out the matching instruction and the correct refined code. After filtering based on the similarity of instruction, the original code from NI is finally converted to a correct instruction-code pair for instruction tuning.

\subsection{Ranking}

Test cases are not only used to validate the correctness of refined codes but also can be seen in a new perspective -- as a measure of difficulty.



Intuitively, the quality of generated inputs depends on to what extent the LLMs understand the instruction. When the instruction is more complex, the constraint to inputs is more stricter. Although a fixed number of inputs is generated, only those who fulfil the requirements of instruction can extract outputs after executions. So the number of test cases constructed can be seen as a measure of difficulty.
Curriculum learning\citep{cl} points out that training data should be from easy to hard. We rank the data in reverse order by the number of test cases so that the model can learn incrementally. 

The process of semi-instruct is described above. The original code from NI is diverse but improper. First generate corresponding instructions, proper refined code, and test cases' inputs by leveraging the generative capability of LLMs like what SI does. Executing the correct original code on inputs can extract gold outputs from successful results to construct test cases. The complete test cases will exclude incorrect refined code. Semi-instruct bridges NI and SI with both advantages. The construct and usage of test cases are novel and effective.




\section{Experiments}


NI and SI dataset are constructed and SemI dataset are converted from NI dataset. We present extensive experiments on a widely used LLM and dataset to show how SemI benefits the tuning process. 

\subsection{Dataset Construction}

\paragraph{Natural-Instruct Dataset} It is not only as a performance comparison but also used to generate SemI dataset. We choose two common datasets \textbf{APPS}~\citep{apps} and \textbf{CodeContest}~\citep{alphacode} to serve as foundation. APPS is a collection of 10k coding problems from Codeforces, Leetcod, etc.. The train split has 5k problems with more than 12k Python solutions. CodeContest is a competitive programming dataset scraped from AtCoder, CodeChef, and two existing datasets Description2Code~\citep{description2code} and CodeNet~\citeplanguageresource{codenet}. CodeContest has more than 13k problems in train split and solutions are written in several languages like Python, Java, and C++. Correct and incorrect solutions are both contained in the original CodeContest. We only retained the correct Python solutions. For the two datasets, instructions are problem descriptions and codes are solutions. 

Before merging into the NI dataset, it is necessary to address the limitations of the two datasets. The following optimisation are implemented.
Delete the problems that need special judge. The rest problems' solutions only need to send input from the command line or parameters passing and the outputs can be directly printed.
Filter out the solutions whose number of tokens is more than 1k. Long solutions are mostly caused by too many meaningless comments or codes. 
Merge the solutions from the same problems. One problem may appear on multiple sites and harvest solutions submitted by different users.
Limit the number of solutions per problem to a maximum of 25 to make the distribution less sharp. Since several problems have more than 1k solutions in the original two datasets, many have less than 10 solutions. 

These approaches can make the NI dataset more realistic and convenient to generate the SemiI Dataset. In the end, the natural-instruct dataset has nearly 8k instructions and 126k codes. 

\paragraph{Self-Instruct Dataset} It is used as the baseline and combined before SemI dataset later. The Code Alpaca project~\citep{codealpaca} aims to build and share an instruction-following Llama model for code generation. We extend its 20k instruction-code pair data generated by text-davinci-003 to 70k through the same self-instruct techniques.

\paragraph{Semi-Instruct Dataset}
We use the original codes from NI dataset to generate SemI dataset. 126k codes are sent to LLM and 92k of them generate instruction, refined code, answer type, and test cases' input successfully. After executing the original code on inputs, nearly 69k piece of data construct at least one test case. The number of refined codes that pass all test cases is 54k. Filtering similar instruction whose ROUGE-L scores with previous data is more than 0.7. Finally, the semi-instruct data have 40k instruction-code pairs. 

\begin{figure}[!ht]
\begin{center}
\includegraphics[width=\linewidth]{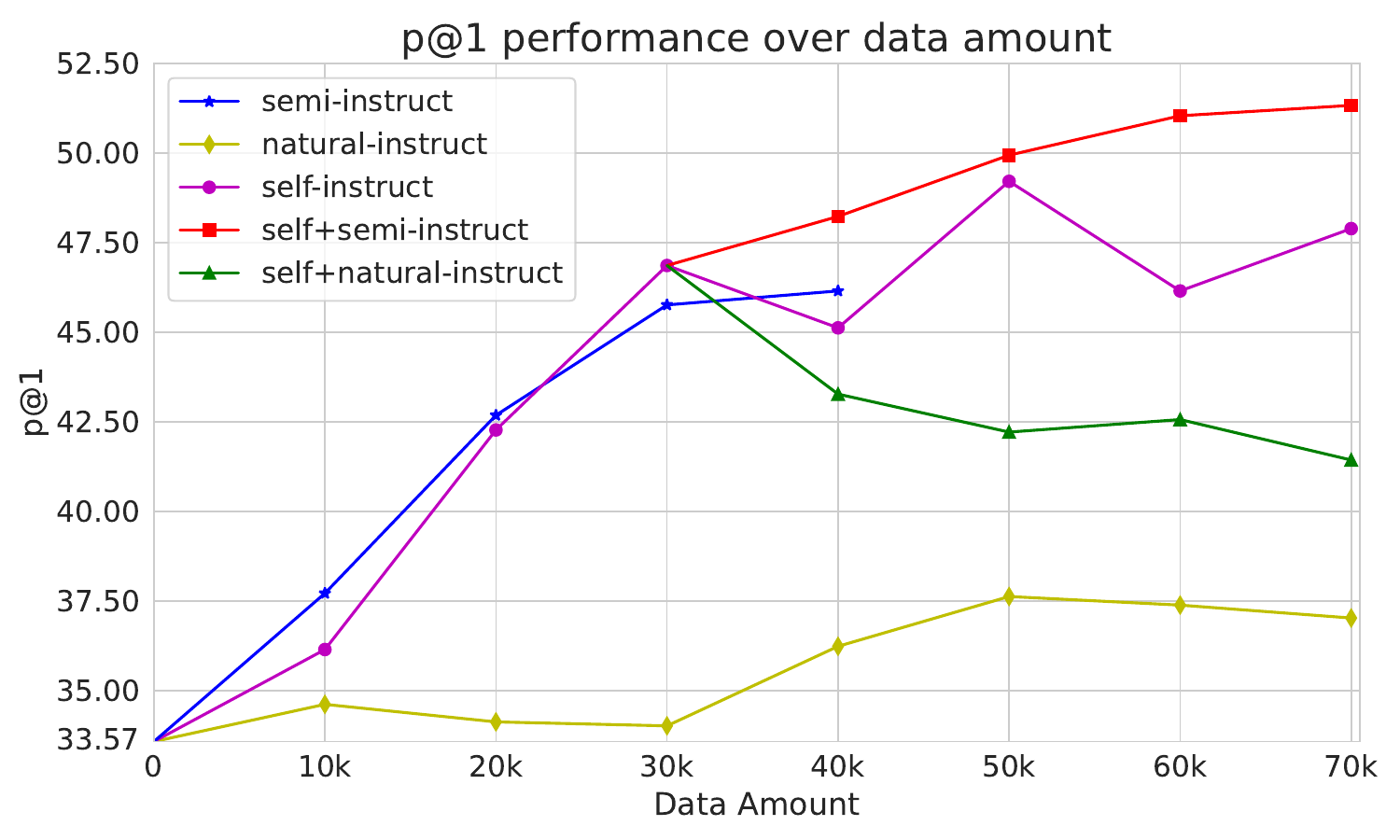} 
\caption{p@1 results on the HumanEval dataset. Note that unlike self-instruct and semi-instruct, the total amount of data for semi-instruct is only 40k.}
\vspace{-1em}
\label{fig.3}
\end{center}
\end{figure}

\paragraph{Data Selection \& Order} 
We conducted experiments on datasets ranging from 10k to 70k entries. Each dataset type had its unique selection method. For the NI datasets, we randomize the problems first. Then, we assemble the related solutions. Data are segmented at each data scale. This ensures new problems were introduced with each data increment. Before training, we sort the data randomly. For the SI dataset, data are used in the generated order. The SemI dataset is treated similarly to NI dataset. However, we sort it by the number of test cases before tuning. These methods aim for real-world emulation and bias minimization.

\subsection{Experiment Setup}

\paragraph{Model} The base model we choose for tunning is StarCoder. It is an open-source 15B parameter Code LLM trained on 1T tokens from GitHub and then finetuned on 35B Python tokens. The performance of StarCoder matches OpenAI code-cushman-001 model on HumanEval~\citep{starcoder, codex}. We follow previous work~\citep{wizardcoder} to generate data by ChatGPT.



\paragraph{Evaluation Dataset} We choose the widely-used dataset -- \textbf{HumanEval}. It consists of 164 Python programming problems used to judge the performance of a model's ability on code generation. Each problem is provided a function name with docstring so the model can then generate the code. Dataset has test cases to check the functional correctness of generated code. 




\paragraph{Metrics} We use the average pass@k score of all problems as the metric of performance evaluation. 
We set $k\in{\{1,10,100\}}$ and sample $200$ times for each problem. p@1 is the proportion of all samples that are correct, which can strictly reflect the correctness. p@10, p@100 can reflect the diversity, the more questions are included in the correct samples the higher these two indicators are.

\begin{table}[!ht]
\centering
\setlength{\tabcolsep}{4pt} 
\begin{tabular}{lccc}
\toprule
\textbf{Dataset} & \textbf{p@1} & \textbf{p@10} & \textbf{p@100} \\
\midrule
base  & $46.86\%$ & $59.62\%$ & $66.09\%$ \\
\midrule
+ self-instruct & $45.12\%$ & $60.23\%$ & $66.92\%$ \\ 
+ natural-instruct & $43.27\%$ & $57.81\%$ & $63.99\%$ \\ 
\midrule
+ semi-instruct & \bm{$48.23\%$} & \bm{$65.10\%$} & \bm{$75.01\%$} \\
\bottomrule
\end{tabular}
\caption{Results among adding 10k self-instruct/natural-instruct/semi-instruct data after 30k self-instruct data. ``base'' represents the performance of base 30k self-instruct data. The best results are marked with \textbf{bold}.}
\vspace{-1em}
\label{tab.1}
\end{table}

\subsection{Implementation Details}
We use the same hyper-parameters from previous work~\citep{wizardcoder} such as limiting the train epochs to 3, the learning rate to 2e-5, the maximum data length to 2048, and the warm-up steps to 30. During the inference phase, we set the temperature to 0.2 and top\_p to 0.95 as common settings in previous work to balance the randomness and determinism. We sample 200 times for each problem and calculate pass@k as the metric.

\subsection{Results}
To fully compare quality and characteristics of NI, SI, and SemI datasets, we conduct five distinct sets of experiments. The first set exclusively utilizes one of the NI, SI, and SemI datasets. Subsequent experiments merge the SI dataset with the SemI or the NI dataset. Mixed datasets are directly combined, maintaining each original order in their respective datasets. 
A visual representation detailing the p@1 metrics can be found in Fig~\ref{fig.3}.

\paragraph{Single Type Data}
The results show that NI only contributes to little performance enhancements, and such gains are often inconsistent. One plausible explanation for the initial decline in performance could be the improper nature of the codes and in NI. This ambiguity potentially misguides model generation. However, as the data scale grows, model start to fit. The impact of such irregularities appears to diminish. The performance decline at later stages might be attributed to the number of instructions added being too small. 

SI demonstrates a notable uplift in performance. However, this improvement does not consistently manifest across varying data scales. This suggests that self-instructed data representations like proper code style are more likely to be understood and learned by models. Reasons for instability could be the lack of sufficient data diversity. As we progressed, there was an apparent rise in duplicated data. Another reason could be incorrect data. The accumulation could potentially distort the model's comprehension of the instructions, subsequently influencing its code generation capabilities.

SemI consistently demonstrates an encouraging trend of improvement. Although SemI is obtained from NI dataset, its huge advantage in performance over NI shows that our method significantly solves the problems of NI. It also shows that the intrinsic value of NI is imprisoned by the simple form and is stimulated by SemI. Compared to SI, SemI only slightly underperforms at 30k but achieves superior results on all the rest data scales. It shows that SemI leverages the benefits of the SI through a SI-like approach. The consistent ascent in performance also underscores its robustness.

In synthesizing the outcomes above, it becomes evident that NI and SI have limitations that impact their performance. SemI that bridging the two approaches shows promising results.

\paragraph{Combined Data}
SemI data is combined with the data of NI and the approach of SI, which shows potential results. We are curious about the upper limit after further combining the data of SI with SemI data.
The performance of SI increases steadily from 10k to 30k, but a large drop occurs at 40k. Therefore, when combining the data, we start with 30k self-instruct and follow it with SemI or NI.

We observe that combining SI with NI resulted in a decline in performance. The distinct distributions of these two datasets likely cause this outcome. Merely merging them seems to merge the individual shortcomings. As detailed in Table~\ref{tab.1}, compared to adding SI, adding NI doesn't augment diversity metrics like p@10 and p@100. In contrast, it reduces them.
However, by converting NI dataset into SemI dataset, the performance keeps improving and significantly outperforms SI. Firstly, because SemI dataset is also generated by LLM like SI dataset, the code is more proper and the instruction is closer to the model's expression, the model can learn more efficiently and directly. When only 10k new data are added, compared to SI, SemI improves on p@1 by 3.11\%. In all scales, we outperform SI with an average improvement of more than 3\%. Second, SemI inherits the diversity of codes in NI. New added data is not duplicated with SI. We improved over SI on p@10 by 4.87\% and on p@100 by 8.09\% in Table~\ref{tab.1}. This shows the great advantage of our method in terms of diversity. Finally, SemI has strong robustness, offering consistent performance enhancements.

\section{Discussion}
\subsection{Ablation Study}
\begin{table*}[t]
\centering
\setlength{\tabcolsep}{10pt} 
\begin{tabular}{lccccccc}
\toprule
\multirow{2}{*}{\textbf{Dataset}} & \multicolumn{3}{c}{\textbf{add 10k data}} & \multicolumn{1}{c}{} & \multicolumn{3}{c}{\textbf{add 20k data}} \\
\cmidrule{2-4} \cmidrule{6-8}
& \textbf{p@1} & \textbf{p@10} & \textbf{p@100} & & \textbf{p@1} & \textbf{p@10} & \textbf{p@100} \\
\midrule
base & $45.12\%$ & $60.23\%$ & $66.92\%$ & & $45.12\%$ & $60.23\%$ & $66.92\%$ \\
+ semi-instruct & \bm{$48.23\%$} & \bm{$65.10\%$} & \bm{$75.01\%$} & & \bm{$49.94\%$} & \bm{$67.90\%$} & \bm{$75.37\%$} \\
\quad - instructions & $40.43\%$ & $52.65\%$ & $61.00\%$ &  & $39.72\%$ & $55.63\%$ & $68.06\%$ \\
\quad - refined code & $38.38\%$ & $54.95\%$ & $66.46\%$ &  & $41.12\%$ & $55.39\%$ & $63.47\%$ \\
\quad - both & $43.27\%$ & $57.81\%$ & $63.99\%$ &  & $42.21\%$ & $56.43\%$ & $65.71\%$ \\
\quad - test cases sort & $44.39\%$ & $62.33\%$ & $71.88\%$ & & $44.74\%$ & $64.53\%$ & $73.56\%$ \\
\quad - all sort & $46.15\%$ & $65.74\%$ & $76.16\%$ &  & $44.46\%$ & $62.60\%$ & $72.46\%$ \\
\bottomrule
\end{tabular}
\caption{ Ablation study on semi-instruct. The left side ``add 10k data" means add 10k new data to 30k self-instruct data, while the right side is adding 20k new data. ``base'' represents the performance of base 30k self-instruct data. ``+semi-instruct" means add semi-instruct data to ``self-instruct". ``-instructions" means replace instructions in added semi-instruct data with their problem descriptions in natural-instruct dataset. ``-refined code" means replace refined codes with original codes. ``-both" means replace both. ``-test cases sort" means only removing ranking in semi-instruct. ``-all sort" means random shuffle all 40k/50k data include self-instruct and semi-instruct.  The best results are marked with \textbf{bold}.}
\vspace{-1em}
\label{tab.2}
\end{table*}

\label{ablationstudy}

To have a deep understanding of the function and importance of each component of SemI, we conduct ablation experiments. To strengthen the reliability, we do a total of 40k data and a total of 50k data respectively, and the main results are shown in Table~\ref{tab.2}. Reducing any of the components in each data scale causes a serious drop in p@1. This means that every component is necessary.

\paragraph{Instructions} Replacing instructions with problem descriptions in NI causes performance degradation at both data scales, even lower than SI without adding new data. When the data scale increases, p@1 drops even more. This is because the increment of problem descriptions is not much compared to a large increase in the number of codes. Instead, more data such as this slows down the model's previous ability to understand and generate.
\paragraph{Refined code} Replacing refined codes with original codes causes a serious degradation at 40k of data, but recovers a little at 50k. This is because The original codes from NI are improper, and different greatly with SI's codes, misleading the model. However, as more such data is added, the model can slowly understand it. This demonstrates the strong adaptive ability of models. It is crucial to note that boosting this capability comes at the expense of previous performance. Adding 20k of such data remains less optimal than without it.
\paragraph{Both} When replacing both instructions and refined codes at the same time, we have empirically found that this is better than replacing one alone. This may stem from the internal consistency of NI data - both problem descriptions and original codes are written by people. Comparison to ``+semi-instruct'' can also be a good way to improve the diversity of SemI, where direct combining of NI and SI does not transfer its diversity, but rather reduces it at p@10, p@100.

\paragraph{Sort} 
Two main orders that are considered in terms of difficulty exist in experiments. One is that the SemI data is in reverse order by the number of test cases, where the harder the problem the fewer test cases will be left; the other is that the SemI dataset is combined after the SI dataset, with the SI considered to be simpler than SemI. Removing the test cases sort not only regresses performance, but the degree of model improvement grows much more slowly when the data scale increases. This suggests that ranking can be effective in improving the efficiency of model learning. When adding 10k SemI data, although ``- all sort'' will be lower on p@1 by 2.08\%, there is an increase on p@10, p@100. This suggests that disrupting the data can slightly increase the diversity of model generation. But when the amount of data increases, this advantage disappears and regression occurs on all metrics. This proves our hypothesis, as the data in NI comes from the competition websites, and SemI inherits significantly higher difficulty than SI.

Through sufficient experiments, we validated each step of the SemI. We avoided the problems of missing instructions and improper codes in NI through LLM, and distinguished the data difficulty through test cases. The performance improvement proved the rationality and superiority of our method.

\subsection{Case Study}

\begin{figure}[!ht]
\begin{center}
\includegraphics[width=\linewidth]{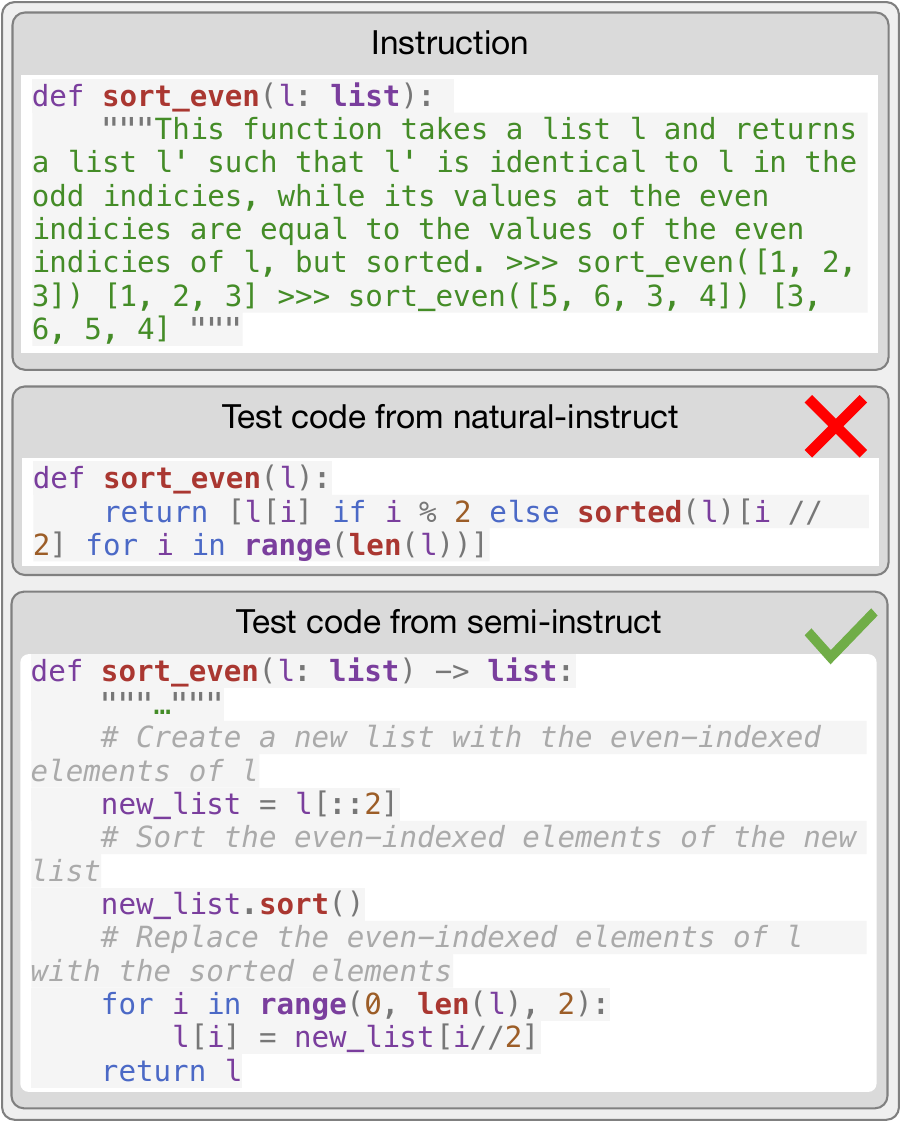} 
\caption{Code generated for the same problem after training the model using the natural-instruct dataset and the semi-instruct dataset, respectively. The former is wrong, and the latter is correct.}
\label{fig.4}
\end{center}
\end{figure}

In addition to the validation of correctness and diversity, we also perform a direct analysis of the codes generated by the models with different data finetune. We compare the generating codes from model training by NI dataset and SemI dataset, as shown in Figure~\ref{fig.4} respectively. NI generates only one nested line of code that fails the test, while SemI generates a clear and correct code.

For some high-level programmers, it is a common trick to write multiple lines of code as one nested line. Such improper codes are more complex and difficult to implement than multi-line codes. NI has a large amount of such code, which adds unnecessary complexity and difficulty. The model mimics human behavior by generating nested single-line code but fails to deal with the logic clearly, resulting in the generation of incorrect code.

From a problem-solving perspective, there is no fundamental difference between nested code and proper code. Compared to NI, SemI's code is more standardized, with docstring introducing the whole function, and each key step is annotated, each line is not nested, which not only increases readability but also avoids complex logic. Through SemI, a large number of improper codes from NI are converted into proper codes, which greatly reduces the difficulty to learn and improves the accuracy.

\section{Conclusion}

In this paper, we propose a novel way of collecting instruction tunning data for code LLMs, semi-instruct, which combines NI and SI. We take the diverse but improper codes in NI and generate proper codes with aligned instructions through a SI-like generation process. Generated codes can't be guaranteed correct. To cope with it, we generate test cases' inputs and execute correct codes in NI to get outputs. The complete test cases can be used to test the correctness of the generated codes. The experiments demonstrate that our method effectively combines the strength of NI and SI.

\clearpage
\section{Bibliographical References}\label{reference}

\bibliographystyle{lrec-coling2024-natbib}
\bibliography{references}

\section{Language Resource References}
\bibliographystylelanguageresource{lrec-coling2024-natbib}
\bibliographylanguageresource{languageresource}



\end{document}